\makeatletter
\def\@fnsymbol#1{\ensuremath{\ifcase#1\or \dagger\or \ddagger\or
   \mathsection\or \mathparagraph\or \|\or **\or \dagger\dagger
   \or \ddagger\ddagger \else\@ctrerr\fi}}
\documentclass[10pt,twocolumn,letterpaper]{article}

\usepackage[pagenumbers]{cvpr} 

\usepackage{multirow}
\usepackage{colortbl}
\newcommand{\name}{{\color{black}{APL }}}
\usepackage[accsupp]{axessibility}

\definecolor{cvprblue}{rgb}{0.21,0.49,0.74}
\usepackage[pagebackref,breaklinks,colorlinks,allcolors=cvprblue]{hyperref}

\def\paperID{1300} 
\def\confName{CVPR}
\def\confYear{2025}

\title{Adaptive Part Learning for Fine-Grained Generalized
Category Discovery: \\ A Plug-and-Play Enhancement}

\author{
Qiyuan Dai \quad Hanzhuo Huang \quad Yu Wu \quad Sibei Yang\thanks{Corresponding author}\\
{School of Information Science and Technology, ShanghaiTech University}\\
{Wuhan University} \quad
{Sun Yat-sen University}\\
}

\begin{document}
\maketitle
\begin{abstract}

Generalized Category Discovery (GCD) aims to recognize unlabeled images from known and novel classes by distinguishing novel classes from known ones, while also transferring knowledge from another set of labeled images with known classes. 
Existing GCD methods rely on self-supervised vision transformers such as DINO for representation learning. However, focusing solely on the global representation of the DINO \texttt{CLS} token introduces an inherent trade-off between discriminability and generalization. 
In this paper, we introduce an adaptive part discovery and learning method, called APL, which generates consistent object parts and their correspondences across different similar images using a set of shared learnable part queries and DINO part priors, without requiring any additional annotations. More importantly, we propose a novel all-min contrastive loss to learn discriminative yet generalizable part representation, which adaptively highlights discriminative object parts to distinguish similar categories for enhanced discriminability while simultaneously sharing other parts to facilitate knowledge transfer for improved generalization. 
Our APL can easily be incorporated into different GCD frameworks by replacing their \texttt{CLS} token feature with our part representations, showing significant enhancements on fine-grained datasets. 
\end{abstract}

\section{Introduction}

\begin{figure*}[t!]
    \centering
    \includegraphics[width=1.0\linewidth]{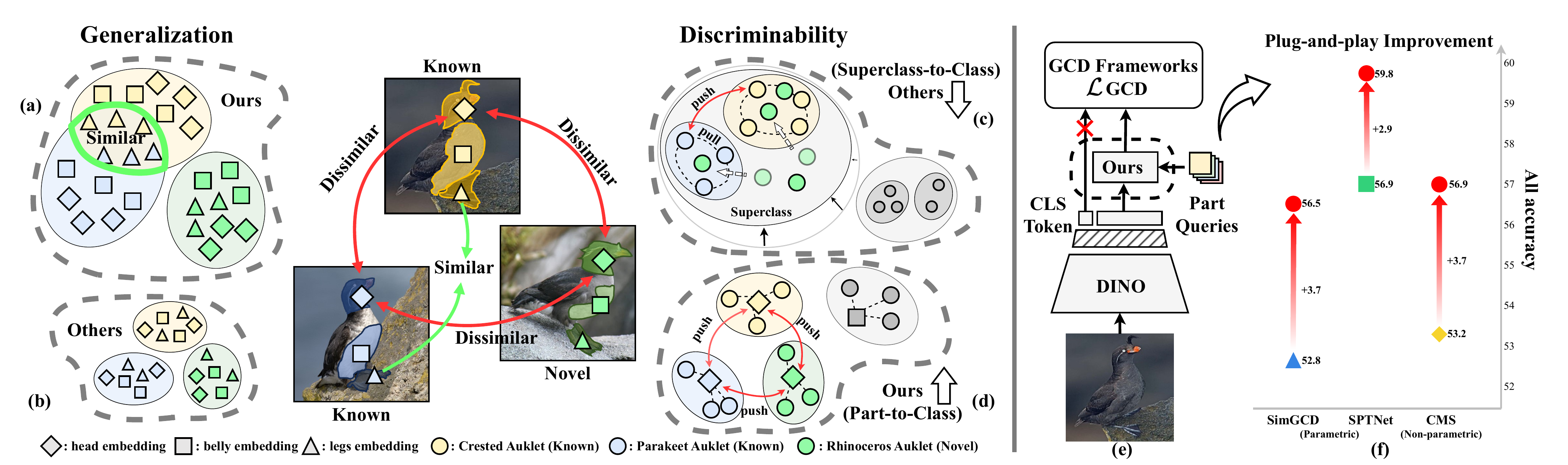}
    \caption{Comparison of discriminability and generalization between our and existing GCD methods that use global \texttt{CLS} token representation. We (a) allow similar parts in similar classes to be shared to achieve knowledge transfer and generalization, while (d) exploiting discriminative parts to distinguish novel classes from similar known classes. In contrast, the existing GCD methods (b) cannot distinguish different parts within the image, are easy to classify novel images into similar known classes when using superclass-based generalization. (e) demonstrates the integration of our APL into existing GCD frameworks, and (f) presents the improvements over diverse baselines.}
    \vspace{-5mm}
    \label{fig:intro}
\end{figure*}

Trained on large-scale annotated data~\cite{deng2009imagenet,lin2014microsoft}
and simplified by the ``closed-world'' assumption, where the test data to be classified adheres to predefined training classes, deep learning models~\cite{dosovitskiy2020image,simonyan2014very,he2016deep}
have made remarkable progress in image recognition, even surpassing human abilities. However, in practical real-world scenarios, this assumption often fails to hold true. This is not only due to the high expense of fully annotating data but, more importantly, due to the impossibility of predefining all categories because of the inherent variability and openness of the real world, where novel classes emerge alongside the known ones. Therefore, Generalized Category Discovery (GCD)~\cite{vaze2022generalized}
seeks to challenge the ``closed-world'' assumption and adapt to an ``open-world'' scenario. In GCD, only a subset of data is labeled with known categories, aiming to assign a category label to each remaining data point, potentially encompassing both known and novel categories. 

Two primary challenges arise in GCD: \textbf{1) Discriminability}: accurately discerning between known and novel classes, and correctly categorizing each; \textbf{2) Generalization}: transferring shared knowledge from known to novel classes to recognizing novel classes without annotations. State-of-the-art GCD methods~\cite{vaze2022generalized,wen2023parametric,wang2024sptnet} primarily exploit the benefits of generalization representation via self-supervised vision transformers (ViT) like DINO~\cite{caron2021emerging} to uphold fundamental generalization capabilities. On one hand, to improve semantic discriminability, some approaches utilize an explicit parametric classifier on known data~\cite{wen2023parametric} or learn prompt-adapted backbones~{\cite{wang2024sptnet,zhang2023promptcal}}. 
On the other hand, to safeguard generalization capabilities, they introduce entropy regularization to mitigate bias towards tending to predict known categories~{\cite{wen2023parametric}} or incorporate shared dynamic concepts~{\cite{pu2023dynamic}}, hierarchies~{\cite{rastegar2024learn}}, or memory bank~{\cite{zhang2023promptcal}} to enhance shared semantics between known and novel classes. 

\label{sec:intro}
However, an inherent trade-off arises between discriminability and generalization when utilizing the image semantics at the global level, namely, the feature representation of ViT's [cls] token, in current GCD methods, particularly in open-world fine-grained recognition. \textbf{\textit{The strong discriminability at the global level impedes the potential for generalization.}} 
The global semantics are rooted in fine-grained local semantics, with the [cls] token's representation obtained through self-attention over local patches~{\cite{dosovitskiy2020image, caron2021emerging}}. Introducing significant differences in the global semantics of similar categories, such as \textcolor[HTML]{CBB163}{crested auklet} and \textcolor[HTML]{A9BAD9}{parakeet auklet} as shown in Figure~\ref{fig:intro}, can improve discriminability but also lead to similarly notable distinctions in their local semantics (refer to Figure~\ref{fig:intro}\textcolor{cvprblue}{b}). Consequently, this impedes the sharing and knowledge transfer of local semantics, such as the legs, which should inherently be similar across the two classes.
As a contrast, Figure~\ref{fig:intro}\textcolor{cvprblue}{a} showcases our discriminable yet shareable local semantics. 
\textbf{\textit{On the contrary, strong generalization on similar categories could impact discriminability on novel categories.}} Hierarchical or superclass-based generalization may result in similar features among different categories within the same superclass, posing challenges in distinguishing novel categories from their similar known counterparts, as shown in Figure~\ref{fig:intro}\textcolor{cvprblue}{c}. This occurs because supervised learning on labeled data only emphasizes differences between similar known classes, leaving the discriminative fine-grained cues of novel categories easily overlooked due to the lack of explicit supervision.
However, our model can distinguish between similar novel and known classes, such as \textcolor[HTML]{A4C092}{rhinoceros auklet} and \textcolor[HTML]{CBB163}{crested auklet}, thanks to our discriminative local semantics (such as the head), as shown in Figure~\ref{fig:intro}\textcolor{cvprblue}{d}. 

In this study, we aim to improve discriminability while preserving generalization through a unified perspective, namely, adaptive part representation learning (APL): \textbf{\textit{highlighting discriminative fine-grained local semantics}} to distinguish similar but distinct categories (e.g., created auklet and parakeet auklet distinguishable by their heads), \textbf{\textit{while also allowing for the sharing of their other local semantics}} (e.g., legs), thereby facilitating knowledge transfer and generalization. 
Specifically, our \name comprises two main components: \textbf{\textit{1) Consistent part discovery and correspondence}}, which learns to decompose images into meaningful semantic parts that can be shared and corresponded one-to-one between similar categories, as illustrated in Figure~\ref{fig:part}\textcolor{cvprblue}{c}. 
However, unsupervised object part discovery and correspondence, without part or even object mask annotations, presents challenges.
Unfortunately, even though DINO's patch features are capable of perceiving object parts~{\cite{oquab2023dinov2}}, naive image partition yields fragmented regions, lacking meaningful parts. Similarly, clustering image patches into a fixed number of groups overlooks the complexity of different scenes, including scenes with missing parts due to occlusion and augmentations, resulting in the noisy part discovery and subsequent incorrect correspondences, as shown in Figure~\ref{fig:part}\textcolor{cvprblue}{a}. Therefore, unlike prior methods, we identify object parts using a set of learnable part queries, guided by DINO's part prior. Each image's patches are then hard-assigned to these queries, forming object parts associated with the same query, as shown in Figure~\ref{fig:framework}. Queries with empty tokens handles potential missing parts. Additionally, inducing shared part queries from all images enhances robustness compared to single-image clustering and naturally establishes correspondences between parts of different images by using queries as bridges. 
2) \textbf{\textit{Diverse and adaptive part representation learning}} acquires more discriminative semantic parts of fine-grained objects while preserving shared parts for improved generalizability across various classes.
To achieve this, we propose two learning objectives: adaptive part contrastive learning and query diversity. Firstly, we propose \textit{all-min} contrastive learning on semantic parts, ensuring that \textit{all} corresponding parts of positive samples are similar while only constraining the \textit{least} similar parts of negative samples to be dissimilar. This enables the adaptive discovery of discriminative parts, while leaving the generalizable parts of negative samples unconstrained, thereby making object part features shareable and generalizable to similar classes. Next, we constrain the diversity between queries, ensuring that each query can focus on and highlight specific semantic parts and details, thus avoiding collapsing into a global semantic representation.

Our \name which improves GCD through discriminative yet generalizable part discovery without relying on a specific GCD network architecture(refer to Figure~\ref{fig:intro}\textcolor{cvprblue}{e}), is integrated into the widely used SimGCD~\cite{wen2023parametric} as well as the latest SPTNet~\cite{wang2024sptnet} and CMS~\cite{choi2024contrastive} to demonstrate its effectiveness. In summary, our contributions are multi-faceted: (1) To the best of our knowledge, we are the first to introduce and differentiate between different object parts for GCD by emphasizing discriminative parts to enhance discriminability, while simultaneously sharing other parts to facilitate knowledge transfer for improved generalization. 
(2) We propose a simple yet effective unsupervised object part discovery method that seamlessly integrates with GCD frameworks to achieve consistent and corresponding part discovery and segmentation across similar images with the same or different categories. 
(3) We propose a novel all-min contrastive learning objective to adaptively select and constrain only the most discriminative parts of negative pairs, fostering discriminative yet generalizable representation learning. 
(4) Our APL is incorporated into the classic SimGCD, recent SPTNet and CMS GCD frameworks, achieving consistent performance improvements across all fine-grained datasets(refer to Figure~\ref{fig:intro}\textcolor{cvprblue}{f}) and enhancing the average accuracy by 3.7\% , 2.9\% and 3.7\% respectively.
\vspace{-2mm}
\section{Related Works}
\noindent\textbf{Generalized Category Discovery} (GCD)~\cite{vaze2022generalized} extends the novel category discovery (NCD) to a more realistic setting, where unlabeled data contains samples of both known and novel classes. 
The pioneering work~\cite{vaze2022generalized} utilizes the self-supervised pre-trained DINO~\cite{caron2021emerging} as the initialization of the feature extractor, in conjunction with semi-supervised k-means clustering. 
It integrates supervised contrastive learning~\textcolor{black}{{\cite{he2020momentum,tian2020contrastive,gutmann2010noise}}} on labeled data for distinguishing known classes, along with unsupervised contrastive learning on different views of both labeled and unlabeled data. 
Later, DCCL~\cite{pu2023dynamic} introduces dynamic conception generation to consider inter-class relationships, using infomap to dynamically generate pseudo-labels.
SimGCD~\cite{wen2023parametric}, unlike prior clustering-based approaches, directly employs a parameterized classification head. 
This circumvents the constraint of clustering methods, which cannot classify individual samples directly, resulting in notable efficiency enhancements. Building upon SimGCD, SPTNet~\cite{wang2024sptnet} introduces spatial prompt tuning to enable the model to leverage information from local regions, while CMS~\cite{choi2024contrastive} modifies the classification head for hierarchical clustering and enhances feature generalization through the mean shift technique.
Some other methods~\cite{weng2023decompose, zhao2021novel} leverage local-level information, but~\cite{weng2023decompose} exploits the structural characteristics of 3D point clouds to facilitate part discovery, whereas~\cite{zhao2021novel} fails to obtain semantically meaningful part information.

\noindent\textbf{Unsupervised Object Part Discovery.} 
To enhance fine-grained image understanding, certain studies advance from the image-level to the object-level and subsequently to an even finer part-level comprehension. 
Early works~\cite{chen2019looks,zheng2017learning} such as ProtoPNet~\cite{chen2019looks} explore the utilization of part-level information to enhance classification. It introduces prototypes into convolutional networks to represent object parts. 
However, the lack of ground truth or priors for accurate object parts limits the robustness of the part information obtained by these methods~\cite{hung2019scops, huang2020interpretable, van2023pdisconet, saha2023particle, aniraj2024pdiscoformer}. 
Other methods~\cite{michieli2020gmnet,zhao2019multi,li2022panoptic} rely on fully annotated object parts to learn part segmentation, aiming to segment images into distinct regions corresponding to different parts of objects. 
Recently, with the emergence of DINO~\cite{caron2021emerging}, some methods use DINO for part discovery. 
For instance, \cite{choudhury2021unsupervised} employ frozen DINO to extract part features for help learning part segmentation. 
Similarly, VLPart~\cite{sun2023going} utilizes DINO to establish correspondence between novel parts and base parts, but it still relies on annotations of object parts in base categories. 
In this paper, by leveraging DINO's coarse part conception, our proposed shared part queries, and novel all-min contrastive loss, we automatically, unsupervised, discover consistent object parts and establish correspondences between them across different images.

\section{Method}
\begin{figure*}[t!]
    \vspace{-0.6cm}
    \centering
    \includegraphics[width=0.8\linewidth]{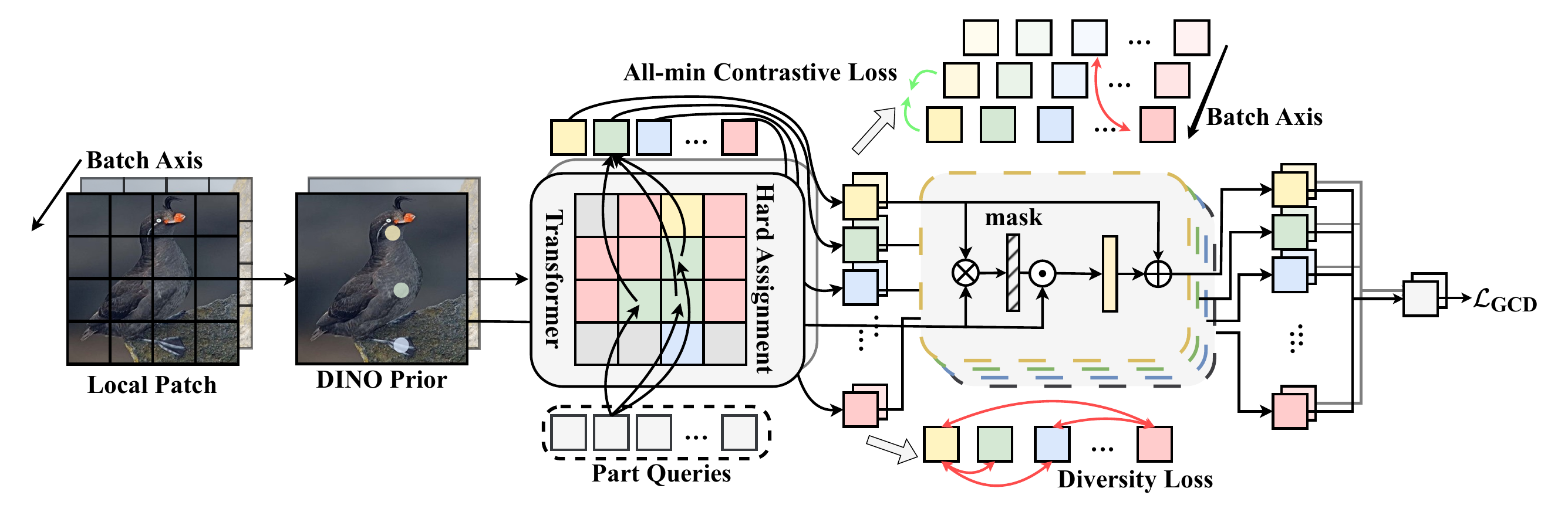}
    \vspace{-6mm}
    \caption{Overall framework of APL. APL achieves consistent and corresponding part discovery, serving as a plug-and-play enhancement.}
    \vspace{-6mm}
    \label{fig:framework}
\end{figure*}
\subsection{Preliminary}
\noindent\textbf{Problem Statement.} The Generalized Category Discovery (GCD) entails categorizing images during inference, with these images potentially belonging to either labeled known classes or unlabeled novel classes during training.
Formally, learning from a labeled set \( \mathcal{D}_l = \{(I_i, c_i)\}_{i=1}^{N_l} \subset \mathcal{I}_l \times \mathcal{C}_l \) and an unlabled set \(\mathcal{D}_u = \{I_i\}_{i=1}^{N_u} \subset \mathcal{I}_u\), each image \(I_i \in \mathbb{R}^{H \times W \times 3}\) in both sets is with height $H$ and width $W$, the GCD aims to accurately categorizes each image \(I_i \in \mathcal{I}_u \) within the label space of unlabeled set \(\mathcal{C}_u\), where \(\mathcal{C}_l \subset \mathcal{C}_u\). Given that the labeled label space \(\mathcal{C}_l\) is a subset of the unlabeled label space \(\mathcal{C}_u\), GCD not only requires discriminability to distinguish between labeled known classes in \(\mathcal{C}_l\) and remaining novel classes in \(\mathcal{C}_u - \mathcal{C}_l\), but also transfer shared knowledge from known to novel classes.

\noindent\textbf{DINO Architecture.}
DINO~\cite{caron2021emerging, oquab2023dinov2} adopts the standard Vision Transformer (ViT)~\cite{vit} architecture. Given an image $I$, DINO initially divides it into a sequence of non-overlapping patch tokens. Each token is then projected into a hidden representation vector via a linear or convolutional layer, resulting in token vectors denoted as 
\(X_\texttt{patch} \in \mathbb{R}^{N \times C}\), where \(N\) is the number of patch tokens, and \(C\) specifies vector dimension. 
To represent the global image feature, an additional learnable \texttt{CLS} token is concatenated to \(X_\texttt{patch}\).
Subsequently, all tokens, patch tokens \(X_\texttt{patch}\) and \texttt{CLS} token, integrated with positional embeddings~\cite{vit}, are fed into the Transformer layers that contain several multi-head self-attention~\cite{attention} (MSA) and feed-forward networks (FFN).
Within the multiple transformer layers, the \texttt{CLS} token interacts with image patch tokens \(X_\texttt{patch}\) through MSA, aggregating global features. Finally, DINO develops robust visual representations, denoted by \(F=[F_\texttt{patch}, F_\texttt{CLS}]\) with token features $F_\texttt{patch} \in \mathbb{R}^{N \times C}$ and \texttt{CLS} feature \( F_\texttt{CLS} \in \mathbb{R}^{1 \times C}\), achieved through view-invariant self-supervised pretraining~\cite{caron2021emerging, oquab2023dinov2}. \textcolor{black}{The DINO architecture is formulated as:}
\begin{equation}
\begin{aligned}
    F_0 & = \phi(\text{PE},[X_\texttt{CLS}, X_\texttt{patch}]), \\
    [F_\texttt{patch}, F_\texttt{CLS}] &\leftarrow F_{l=L}, \ F_l = \text{FFN}(\text{MSA}(F_{l-1})),
    \label{equ:dino}
\end{aligned}
\end{equation}
where $[\cdot,\cdot]$ denotes concatenation, PE represents position encoding, $\phi$ is the fusion function of position encoding and tokens~\cite{caron2021emerging}, and $l$ and $L$ represent the index of the $l$-th transformer layer and the total number of transformer layers, respectively.

\label{sec:dinopart}
\noindent\textbf{Part Perception in DINO.} 
With self-supervised pretraining and ViT architecture, DINO exhibits many functionalities that are absent in fully supervised training~\cite{vit} or ConvNet~\cite{resnet} networks, one of which is the perception of object parts. Previous works~\cite{caron2021emerging, choudhury2021unsupervised} find that in the MSA interactions between the \texttt{CLS} token and image patch tokens, the overall attention scores from all $M$ attention heads can roughly reveal the foreground object, while \textcolor{black}{attention scores} of each individual head can highlight semantically meaningful parts of the object. 
Specifically, we use the attention scores from the final-layer MSA of DINO to obtain part attention. In the \(m\)-th attention head, with the features of the \texttt{CLS} token denoted as \(F^{m}_\texttt{CLS} \in \mathbb{R}^{1 \times \frac{C}{M}}\) and the features of the patch tokens as \(F^{m}_\texttt{patch} \in \mathbb{R}^{N \times \frac{C}{M}} \), the calculation of $m$-th part attention \(A_m\) is as follows:
\begin{equation}
\begin{aligned}
{W_m} &= \text{Softmax}(\frac{\text{Proj}_{\text{Q}}^m(F^m_\texttt{CLS}) \ast \text{Proj}_{\text{K}}^m([F^m_\texttt{CLS}, F^m_\texttt{patch}])^\top}{\sqrt{C}}),  \\
A_m &\leftarrow W_m[1, 2:N+1],
\end{aligned}
\label{equ:msa}
\end{equation}
where $\text{Proj}_{\text{Q}}^m$ and $\text{Proj}_{\text{K}}^m$ denote projection matrix of \(m\)-th head, $\top$ represents matrix transpose. The ${W_m \in \mathbb{R}^{1 \times {(N+1)}}}$ denotes the attention scores between the \texttt{CLS} token and all tokens, encompassing both the \texttt{CLS} token and patch tokens. Specifically, the attention scores \(A_m \in \mathbb{R}^{1 \times N}\) for patch tokens highlight potential patch regions where parts could be located, essentially constituting the part attention in the \(m\)-th head. 

\subsection{Consistent Part Discovery and Correspondence}
\noindent\textbf{The Motivation for Introducing Object Parts} is to highlight discriminative object parts for enhanced discriminability, while simultaneously sharing other parts to facilitate knowledge transfer for improved generalization. Since GCD requires learning from unlabeled data, DINO—a self-supervised, pre-trained model that learns from unlabeled images and their augmentations—is a natural choice for existing GCD methods\textcolor{black}{~\cite{wen2023parametric,wang2024sptnet}} seeking to obtain generalizable visual representations. Existing GCD methods predominantly use the global semantics of images, particularly the features of DINO's \texttt{CLS} token, for classification or clustering to categorize images. However, these global features may lead to a trade-off between discriminating similar categories and generalizing from similar known to novel ones, as detailed in Sec~\ref{sec:intro}. Therefore, we advocate for integrating local semantics, such as object parts, into GCD. Notably, object parts are not merely naive supplements of fine-grained local semantics to global features. More crucially, by delineating the roles of different object parts, discriminative parts can effectively differentiate similar categories, while also allowing for the sharing of other parts across these categories, thereby improving both discriminability and generalization.

\begin{figure}[t!]
    \centering
    \includegraphics[width=0.98\linewidth]{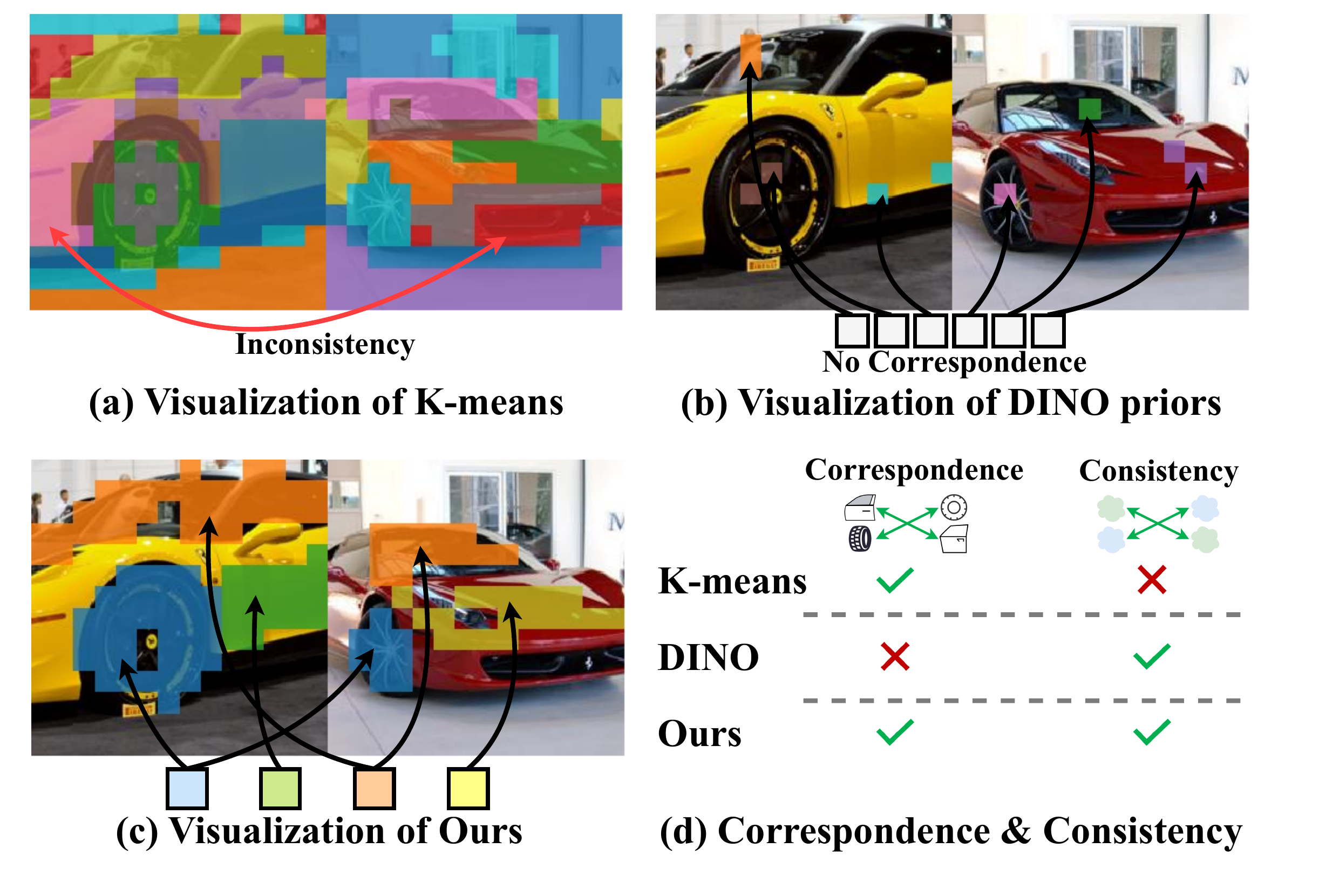}
    \caption{Part discovery by (a) K-means, (b) DINO priors, and (c) ours. Only ours demonstrates consistency in handling partial parts and correspondences between similar images.}
    \vspace{-5mm}
    \label{fig:part}
\end{figure}

\noindent\textbf{Three Principles of Our Part Discovery.} \textbf{(1) Consistency:} object parts in images of the same category should be consistent, especially when some images may have missing parts. 
For instance, in cars, parts like windows, tires, and hoods are common, but due to image augmentation or occlusion, some images may only feature partial parts. 
Simple single-image-based grouping or clustering on DINO's patch features is highly unreliable, potentially leading to meaningless or inconsistent discoveries of object parts across images, as shown in Figure~\ref{fig:part}\textcolor{cvprblue}{a}. \textbf{(2) Correspondence:} some object parts of similar categories should correspond to each other, facilitating the identification of both discriminative and shareable parts. For instance, \textcolor{black}{dogs and cats share common parts like the head, tail, and body.} 
However, only by matching their corresponding parts can we identify the head as potentially more discriminative, while body features may be better suited for sharing and generalizing to other canid species. 
\textbf{(3) Efficiency:} part discovery should be seamlessly integrated into existing GCD frameworks without the need for additional annotations, external models, or significant computational complexity.

\noindent\textbf{Consistent Part Discovery and Correspondence.} To adhere to our third principle of efficiency, we propose directly leveraging DINO, drawing inspiration from its perceptual abilities for object parts (see Sec~\ref{sec:dinopart}), and integrating part discovery into the online learning process of GCD. 
However, DINO's part perception is too sensitive to variations among different images, thus failing to meet our principles of consistency and correspondence, as shown in Figure~\ref{fig:part}\textcolor{cvprblue}{b}. Therefore, instead of solely depending on DINO's part perception, we leverage it as a prior to facilitate learning a set of shared learnable part queries across all images for object discovery, as shown in Figure~\ref{fig:framework}. These cross-image, unified part queries not only ensure more robust object discovery compared to single-image-based methods but also naturally serve as bridges for building correspondence between different images. Specifically, let us denote the set of learnable part queries as \(Q \in \mathbb{R}^{T \times C}\), where \(T\) is the number of queries and \(C\) is the dimension number. First, for an input image \(I\), we extract its DINO part priors $F_\texttt{prior} \in \mathbb{R}^{M \times C}$ based on DINO's \(M\) part attentions \([A_1, A_2, ..., A_M]\) and DINO patch features \textcolor{black}{\(F_\texttt{patch}\)}, defined in Sec~\ref{sec:dinopart}. Then, we transform the shared learnable queries $Q$ into image-specific ones $Q_\texttt{I} \in \mathbb{R}^{T \times C}$ via cross-attention between $Q$ and the prior $F_\texttt{prior}$, as formulated below:
\begin{equation}
\begin{aligned}
F_\texttt{prior} &= [\epsilon(A_1) \ast F_\texttt{patch}, \cdots,\epsilon(A_M) \ast F_\texttt{patch}],\\
Q_{I} &= \text{Proj}_{\text{Q}}(Q) \ast \text{Proj}_{\text{K}}(F_\texttt{part})^{\top} / \sqrt{C},
\label{equ:priors}
\end{aligned}
\end{equation}
where \(\epsilon\) represents a threshold filter function to transform \(A_m\) into a binary matrix to highlight the part regions, \(\text{Proj}_{\text{Q}}\) and \(\text{Proj}_{\text{K}}\) are the learnable projection matrices. The \( \epsilon(A_m) \ast F_\texttt{patch}\) represents the feature of the \(m\)-th part in DINO, and the features from all heads form DINO's part priors \(F_\texttt{prior}\). 
Part queries \(Q_\texttt{I}\) encode the features of potential object parts in image \(I\). Next, object parts are identified by hard-assigning image patches to the queries based on the similarity of their features $F_\texttt{patch}$ and $Q_\texttt{I}$, where patches within the same query constitute a potential object part. As the hard assignment, which is the $\arg\max$ over similarities, is not differentiable, we approximate the hard assignment and preserve differentiability by Gumbel-Softmax distribution~\cite{gumbel1954statistical,maddison2014sampling}, as formulated as follows:
\begin{equation}\label{eq:4}
    \begin{aligned}
    &H_\texttt{g} = \mathrm{softmax}((\frac{\text{Proj}_{\text{Q}}(Q_I) \ast \text{Proj}_{\text{K}}(F_\texttt{patch})^{\top} }{\sqrt{C}} + G)/\mathop{\tau}\limits_{\ }),\\
    &\begin{split}
        H_\texttt{part} = &(\mathrm{one\_hot}(\arg\max_N{(H_\texttt{g})}))^\top-\mathrm{sg}(H_\texttt{g})+ H_\texttt{g},
    \end{split}\\
    &P_\texttt{part} = H_\texttt{part} \ast F_\texttt{patch}, 
    \end{aligned}
\end{equation}
where each Gumbel noise \(g_n \in G\) with \(g_n = -\log(-\log(u_n))\) and \(u_n\) are i.i.d. samples drawn from Uniform(0,1), \(\tau\) is the temperature. \(\text{Proj}_{\text{Q}}\) and \(\text{Proj}_{\text{K}}\) denote the learnable projection matrices of \(Q_I\) and \(F_\texttt{patch}\). 
The $H_\texttt{g}$ approximates sampling the $\mathrm{one\_hot}(\arg\max_N(\text{Proj}_{\text{Q}}(Q_I) \ast \text{Proj}_{\text{K}}(F_\text{patch})^{\top} / \sqrt{C}+G))$, where $\arg\max_N$ selects the corresponding patch token with the highest attention score for each part query, and $\mathrm{one\_hot}$ transforms the patch token index into a one-hot vector. $sg$ represents the stop gradient operator. 
The resulting binary \(H_\text{part} \in \mathbb{R}^{T \times N}\) indicates the hard assignment from \(N\) patches to \(T\) part queries. Each \(H_t \in \mathbb{R}^{1 \times N}\) within \(H_\text{part}\) represents a potential part, with indices containing a value of 1 indicating which patches belong to that specific part. The \(P_t \in \mathbb{R}^{1 \times C}\) within \(P_\text{part}\) represents the feature of each part $H_t$. 

Our query hard assignment for part discovery can adaptively handle consistency and situations where some parts are absent. In such cases, when an object part is too small and its constituent patches are scattered, the corresponding query becomes a meaningless placeholder for the potential missing part. Additionally, it ensures part correspondence across images. For images \(I_a\) and \(I_b\), their respective parts \textcolor{black}{\( H_\texttt{part}^a\) and \(H_\texttt{part}^b\)} are obtained through shared part queries \(Q\). Consequently, parts corresponding to same query, \textcolor{black}{such as $H_t^a$ and $H^b_t$ of the \(t\)-th query}, align with each other.

\vspace{-2mm}
\subsection{Diverse and Adaptive Part Learning}
\vspace{-2mm}
While object parts are meant to represent distinct components, without constraints on the learning of randomly initialized learnable parts, they easily correspond to arbitrary image regions rather than meaningful parts or collapse into identical global patches. Additionally, automatically distinguishing discriminative and transferable object parts enables learning their respective representations, crucial for enhancing GCD from both discriminative and generalizable perspectives. Therefore, we propose a novel \textcolor{black}{all-min contrastive learning} objective combined with a diverse loss to facilitate part learning. 

\noindent\textbf{All-min Contrastive Learning.} Its core idea involves (1) constraining all corresponding object parts in a positive pair to be more similar than non-corresponding pairs, where positive pairs originate from images of the same class in the labeled set or from images and their augmentations in both the labeled and unlabeled sets. This fosters the learning of distinct parts within a class and consistency among parts across images of the same class. (2) In negative pairs of images from different classes, we only enforce dissimilarity on the most dissimilar corresponding part, leaving the similarity of their other corresponding parts unconstrained. This approach capitalizes on the least similar corresponding part's discriminative power in distinguishing between the two images, while permitting other parts to be similar to facilitate the sharing of similar components for generalization. 
Specifically, for a batch of images denoted as \( B \), our all-min contrastive loss is formulated as follows: 
\begin{equation}
\scriptsize
\begin{aligned}
\mathcal{L}_\text{all-min} &= -\frac{1}{|B|} \sum_{I_a\in B} \log \frac{ \sum_{I_{b+}\in S^+(I_a)} \sum^{T}_{t=1}\text{exp}(\text{sim}(P^{a}_t, P^{b+}_t) /\mathop{\tau})}{\sum_{I_{b-}\in S^-(I_a)} \text{exp}(\min_{t \in [1,T]}\text{sim}(P^a_t, P^{b-}_t) /\mathop{\tau})},
\end{aligned}
\end{equation}
where \( I_a \) represents an image in the batch \( B \), the $\text{sim}$ means the cosine similarity between corresponding parts from two different images or two different augmentation, the sets \( S^+(I_a) \) and \( S^-(I_a) \) represent the set of positive and negative image samples of \( I_a \), respectively. 
If $I_a$ is labeled, i.e., $I_a \in \mathcal{D}_l$, the positive set \( S^+(I_a) \) consists of labeled images \( I_i \in \mathcal{D}_l \) whose class \( c_i \) is the same as the class \( c_a \) of image \( I_a \), i.e., \( S^+(I_a) = \{ I_i \in \mathcal{D}_l : \, c_i = c_a \} \cap B \). Otherwise, the set only contains augmentation images, i.e., \( S^+(I_a) =  \{ I_i \in B: \, \text{ID}(I_i) = \text{ID}(I_a) \}\), where $\text{ID}$ returns the id of the image. Contrary to the positive sets, if \( I_a \) is labeled, the negative set consists of labeled images with different classes from \( I_a \), i.e.,  \( S^-(I_a) = \{ I_i \in \mathcal{D}_l : \, c_i \neq c_a \} \cap B \). For unlabeled \( I_a \), we randomly sample some images from \( B \) as negative samples, ensuring that not all parts of these images are similar to each corresponding part of \( I_a \), and use the similarity of all positive pairs in the batch as the threshold for sampling negative images.

\noindent\textbf{Diverse Object Parts.} To prevent different part queries from collapsing onto the same semantic parts, we apply diversity loss to the part queries, encouraging them to focus on distinct object parts. Specifically, we constrain that different parts within an image are orthogonal:
\begin{equation}
\mathcal{L}_\text{div}=\sum_{P_t \in P_{part}} \sum_{\hat{P_t} \in P_{part}\setminus P_t} \max (0, \text{sim}( P_t,\hat{P_t})),
\end{equation} 
where $P_t$ and $\hat{P_t}$ are the part features of two different parts within an image, and sim denotes the cosine similarity.

\subsection{A Plug-and-Play Integration}
Since our part discovery and learning do not rely on any specific GCD framework but only require a DINO backbone adopted by all existing GCD methods, we can integrate it into various GCD frameworks, such as the parameterized classification head in SimGCD~\cite{wen2023parametric} and SPTNet~\cite{wang2024sptnet}. The parametrized framework freezes most layers of DINO and only tunes the final layer, treating the features of the $\texttt{[CLS]}$ token, $F_{\texttt{token}}$, as the sample's feature representation for class prediction. Its learning objective includes both representation learning with InfoNCE loss and parametric classification on $F_{\texttt{token}}$~\cite{wen2023parametric,wang2024sptnet}, formulated as follows: 
\begin{equation} \label{eq:gcd}
\mathcal{L}_\text{GCD}=(1-\lambda)(\mathcal{L}^u_\text{cls}+\mathcal{L}^u_\text{rep}) + \lambda (\mathcal{L}^s_\text{cls} + \mathcal{L}^s_\text{rep})
\end{equation} 
where $\lambda$ is the balance factor, $\mathcal{L}^u_\text{cls}+\mathcal{L}^u_\text{rep}$ and $\mathcal{L}^s_\text{cls} + \mathcal{L}^s_\text{rep}$ represent the sum of representation learning loss and parametric classification loss on labeled and unlabeled sets, respectively. 

To solely showcase the discriminative yet generalizable effectiveness of our object parts, we simply pool the features from different parts to obtain an overall part feature, which replaces the feature $F_\texttt{CLS}$ of \texttt{CLS} token for both classification prediction and feature representation learning, as illustrated in Figure~\ref{fig:framework}\textcolor{cvprblue}{b}.
The overall learning objective of the model is as follows:
\begin{equation}
\mathcal{L}_\text{overall}=(1-\lambda)(\mathcal{L}^u_\text{cls}+\mathcal{L}^u_\text{rep}) + \lambda (\mathcal{L}^s_\text{cls} + \mathcal{L}^s_\text{rep}) + \mathcal{L}_\text{div} + \mathcal{L}_\text{all-min}
\end{equation}

\begin{table*}[htp]
\vspace{-5mm}
\small
\renewcommand{\arraystretch}{0.85}
\setlength{\tabcolsep}{7pt}
\begin{tabular}{lcccccccccccc}
\toprule
\multirow{2}{*}{Methods}&   \multicolumn{3}{c}{CUB} & \multicolumn{3}{c}{Stanford Cars} & \multicolumn{3}{c}{FGVC-Aircraft}& \multicolumn{3}{c}{Herbarium-19}\\
\cmidrule(rl){2-4}\cmidrule(rl){5-7}\cmidrule(rl){8-10}\cmidrule(rl){11-13}
     & All  & Known  & Novel  & All  & Known  & Novel  & All  & Known  & Novel & All  & Known  & Novel \\
\midrule
$k$-means~\cite{arthur2007k}  & 34.3 & 38.9 & 32.1 & 12.8 & 10.6 & 13.8 & 16.0 & 14.4 & 16.8 &13.0&12.2&13.4 \\
RS+~\cite{han2020automatically}         & 33.3 & 51.6 & 24.2 & 28.3 & 61.8 & 12.1 & 26.9 & 36.4 & 22.2 &27.9&55.8&12.8\\
UNO+~\cite{fini2021unified}              & 35.1 & 49.0 & 28.1 & 35.5 & 70.5 & 18.6 & 40.3 & 56.4 & 32.2 &28.3&53.7&14.7\\
ORCA~\cite{cao2021open}                    & 35.3 & 45.6 & 30.2 & 23.5 & 50.1 & 10.7 & 22.0 & 31.8 & 17.1 &20.9&30.9&15.5\\
GCD~\cite{vaze2022generalized}        & {51.3} & {56.6} & {48.7} & {39.0} & 57.6 & {29.9} & {45.0} & 41.1 & {46.9} &35.4&51.0&27.0\\
CiPR~\cite{hao2023cipr}       & {57.1} & {58.7} & {55.6} & {47.0} & 61.5 & {40.1} & {-} & - & {-}  &-&-&-\\
GPC~\cite{zhao2023learning}       & {52.0} & {55.5} & {47.5} & {38.2} & 58.9 & {27.4} & {43.3} & 40.7 & {44.8}  &-&-&-\\
DCCL~\cite{pu2023dynamic}      & {63.5} & {60.8} & {64.9} & {43.1} & 55.7 & {36.2} & {-} & -& {-}  &-&-&-\\
PromptCAL~\cite{zhang2023promptcal}      & {62.9} & {64.4} & {62.1} & {50.2} & 70.1 & {40.6} & {52.2} & 52.2 & {52.3}  &37.0&52.0&28.9\\
PMI~\cite{chiaroni2023parametric}             &62.7  & 75.7 & 56.2& 43.1 & 66.9 &31.6  & - & - & - &42.3&56.1&34.8 \\
\midrule
SimGCD~\cite{wen2023parametric}                   & {60.3} & {65.6} & {57.7} & {53.8} & {71.9} & {45.0} & {54.2} & {59.1} & {51.8}  &43.0  &58.0&35.1\\
\rowcolor{gray!35}
Ours (SimGCD) & {64.5} & {68.1} & {62.1} & {60.1} & {77.6} & {51.2} & {56.6} & {60.2} & {54.8}  &44.9 &58.1&37.9\\
\rowcolor{gray!15}
$ \Delta$& \textcolor{BlueGreen}{+4.2} &\textcolor{BlueGreen}{+2.5} & \textcolor{BlueGreen}{+4.4} & \textcolor{BlueGreen}{+6.3} & \textcolor{BlueGreen}{+5.7} & \textcolor{BlueGreen}{+6.2} & \textcolor{BlueGreen}{+2.4} & \textcolor{BlueGreen}{+1.1} & \textcolor{BlueGreen}{+3.0}  & \textcolor{BlueGreen}{+1.9}& \textcolor{BlueGreen}{+0.1}& \textcolor{BlueGreen}{+2.8}\\
SPTNet~\cite{wang2024sptnet}                & {65.8} & {68.8} & {65.1} & {59.0} & {79.2} & {49.3} & {59.3} & {61.8} & {58.1}  &43.4 &58.7&35.2\\

\rowcolor{gray!35}
Ours (SPTNet)  & {68.5} & {73.1} & {66.2} & {62.3} & {80.7} & {53.4} &{60.9} & {63.5} & {59.6} &47.3 &66.6&37.0\\
\rowcolor{gray!15}
$\Delta$ & \textcolor{BlueGreen}{+2.7} &\textcolor{BlueGreen}{+4.3} & \textcolor{BlueGreen}{+1.1} & \textcolor{BlueGreen}{+3.3} & \textcolor{BlueGreen}{+1.5} & \textcolor{BlueGreen}{+4.1} & \textcolor{BlueGreen}{+1.6} & \textcolor{BlueGreen}{+1.7} & \textcolor{BlueGreen}{+1.5}  & \textcolor{BlueGreen}{+3.9}& \textcolor{BlueGreen}{+7.9}& \textcolor{BlueGreen}{+1.8}\\

CMS~\cite{choi2024contrastive}             &68.2  & 76.5 &64.0 & 56.9 &76.1  & 47.6 & 56.0 & 63.4 &52.3  &36.4&54.9&26.4 \\
CMS$^{\dag}$            &66.6  & 74.8 &62.6 & 56.1 &76.4  & 47.1 & 54.1 & 61.7 &50.3  &  35.9  & 55.1 & 25.6 \\
\rowcolor{gray!35}
Ours (CMS$^{\dag}$)  & {69.2} & {75.3} & {66.1} & {62.1} & {81.8} & {52.6} &{57.9} & {66.0} & {53.8} &38.4 &57.2&28.2\\
\rowcolor{gray!15}
$\Delta$ & \textcolor{BlueGreen}{+2.6} &\textcolor{BlueGreen}{+0.5} & \textcolor{BlueGreen}{+3.5} & \textcolor{BlueGreen}{+6.0} & \textcolor{BlueGreen}{+5.4} & \textcolor{BlueGreen}{+5.5} & \textcolor{BlueGreen}{+3.8} & \textcolor{BlueGreen}{+4.3} & \textcolor{BlueGreen}{+3.5}  & \textcolor{BlueGreen}{+2.5}& \textcolor{BlueGreen}{+2.1}& \textcolor{BlueGreen}{+2.6}\\
\bottomrule
\end{tabular}
\caption{Evaluation on the Semantic Shift Benchmark (SSB) and Herbarium-19 benchmarks.$^{\dag}$ represents our reproduced result.} 
\label{table1}
\vspace{-5mm}

\end{table*}

\section{Experiment}
\subsection{Experiment Setup}
\noindent\textbf{Datasets.} 
We follow previous approaches~\cite{vaze2022generalized,wen2023parametric} to evaluate our method on four fine-grained and three generic image classification datasets. 
Following them, we adopt the same partitioning strategy across these datasets: \textcolor{black}{for all fine-grained datasets~\cite{wah2011caltech, krause20133d, maji2013fine, tan2019herbarium}, CIFAR-10~\cite{krizhevsky2009learning}, and ImageNet-100~\cite{deng2009imagenet}, $50\%$ of the classes are considered as known classes, with the remaining classes treated as novel classes.}
For CIFAR-100~\cite{krizhevsky2009learning}, 80 classes are designated as known classes. Within these known classes, $50\%$ samples from each class are chosen to construct the labeled set $\mathcal{D}_l$, while the remaining samples constitute the unlabeled set $\mathcal{D}_{u}$, along with the images of novel classes. 
For detailed information, please refer to the Appendix. 

\noindent\textbf{Evaluation Metric.} We evaluate the effectiveness of our method on unlabeled dataset $\mathcal{D}_u$ using clustering accuracy, following~\cite{wen2023parametric,wang2024sptnet,choi2024contrastive}. The accuracy is defined as $\text{ACC} = \frac{1}{|\mathcal{D}_u|} \sum_{i=1}^{|\mathcal{D}_u|} 1(y_i = f^{*}(\hat{y}^i))$, 
\text{where  $f^* = \arg \max_{f \in \mathcal{F}} \sum_{i=1}^{M} 1(y_i = f(\hat{y}^i))$}, and $\mathcal{F}$ is all possible matching between the predicted labels with the ground truth. 
\begin{table}[b]
\vspace{-5mm}
\small
\renewcommand{\arraystretch}{0.98}
\setlength{\tabcolsep}{5pt}
\begin{tabular}{lcccccc}

\toprule
{\multirow{2}{*}{Methods}}&    \multicolumn{2}{c}{CIFAR-10} & \multicolumn{2}{c}{CIFAR-100} & \multicolumn{2}{c}{ImageNet-100} \\
\cmidrule(rl){2-3}\cmidrule(rl){4-5}\cmidrule(rl){6-7}
                                 & All    & Novel  & All    & Novel  & All   & Novel \\
\midrule

InfoSieve~\cite{rastegar2024learn}                      & {94.8} & \cellcolor{gray!15}{93.4} & {78.3}  & \cellcolor{gray!15}{70.5} & {80.5} & \cellcolor{gray!15}{73.8} \\
SelEx~\cite{rastegar2024selex}                      & {95.9}  & \cellcolor{gray!15}{94.8} & {82.3}  & \cellcolor{gray!15}{76.3} & {83.1}  & \cellcolor{gray!15}{77.8} \\
Ours(SimGCD)                    & {97.1} & \cellcolor{gray!15}\textbf{98.2} & {80.9} & \cellcolor{gray!15}\textbf{79.5} & {83.2}  & \cellcolor{gray!15}\textbf{78.5}\\ 
\bottomrule
\end{tabular}
\caption{Evaluation on generic image recognition datasets.} \label{table_generic}

\end{table}
\subsection{Comparison with the State-of-the-Art Methods}

Table~\ref{table1} compares our method with state-of-the-art models (SOTAs) on fine-grained datasets, including the SSB benchmark and the Herbarium19 dataset. Our method achieves state-of-the-art performance across multiple datasets and consistently outperforms our three baselines, SimGCD~\cite{wen2023parametric} SPTNet~\cite{wang2024sptnet} and CMS~\cite{choi2024contrastive}. Specifically, compared to SOTAs, we achieve average improvements of 2.2\% on the SSB dataset across novel categories and notable enhancements on the more challenging Herbarium dataset, with improvements of 3.9\%, 7.9\%, and 1.8\% across all, known, and novel categories, respectively. 

Compared to the pioneering parameterized-head SimGCD, we achieve direct performance improvements on the four fine-grained datasets, with gains of 4.2\%, 6.3\%, 2.4\%, and 1.9\%, respectively. This indicates the effectiveness of our part discovery, as we simply replace the \texttt{CLS} token used by SimGCD in representation learning and class prediction with the average of our different part features. 
Particularly, we note that our improvements on novel classes are even more significant, with gains of 4.4\%, 6.2\%, 3.0\%, 2.8\%, respectively. This indicates that our method, by distinguishing between discriminative and transferable parts within similar categories and facilitating feature sharing among transferable parts, aids in the generalization from known to novel classes. 
Moreover, compared to SPTNet, which also utilizes local information, our method still obtains performance gains of 2.7\%, 3.3\%, 1.6\%, and 3.9\%, respectively, demonstrating the positive impact of our explicit object part discovery and utilization. 
Unlike SPTNet, which implicitly enhances attention to local information through spatial visual prompts, our method directly extracts semantically meaningful object parts from images adaptively, which is more intuitive and interpretable. Additionally, by introducing our part learning loss on part tokens, we emphasize the diversity of parts and their adaptive discrimination between paired images, enhancing feature discriminability and generalization compared to globally unified visual prompts. 
In contrast to DCCL and InfoSieve, which top-down establish category relationships hierarchically or superclass-based, our approach establishes relationships from bottom to top through parts, resulting in enhanced generalization and boosting the accuracy of novel categories by 7.0\% and 7.1\% on the Stanford Cars and FGVC-Aircraft datasets, respectively. 

\begin{figure}[thbp!]
    \vspace{-0.3cm}
    \centering
    \includegraphics[width=0.9\linewidth]{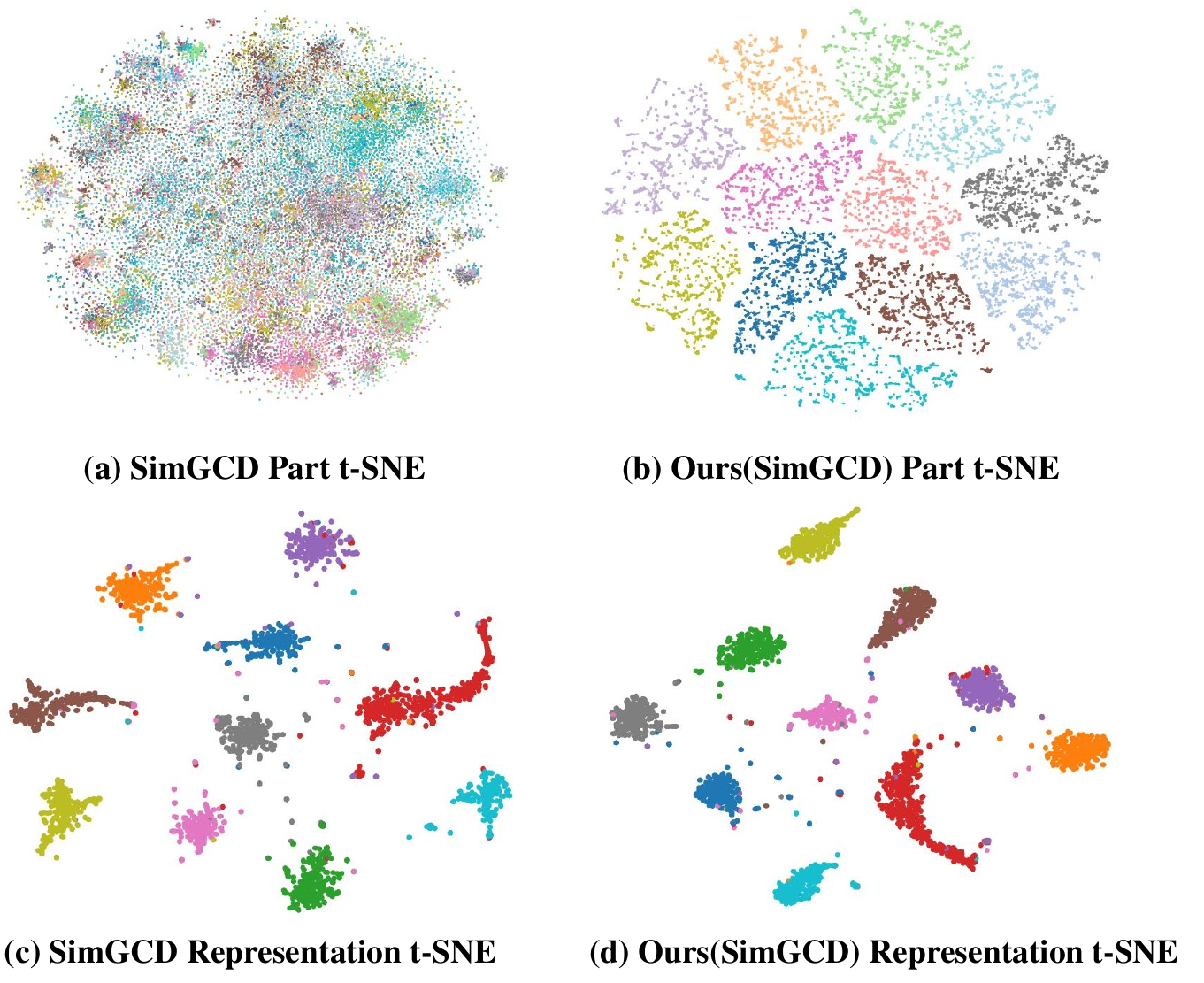}     
    \caption{t-SNE visualization of part features on CUB-200 and image features on randomly sampled 10 classes on CIFAR-100. Each color represents a specific part (category).}
    \label{fig:parttsne}
    \vspace{-4mm}
\end{figure}

\begin{figure}[t!]
    \centering
    \includegraphics[width=0.5\linewidth]{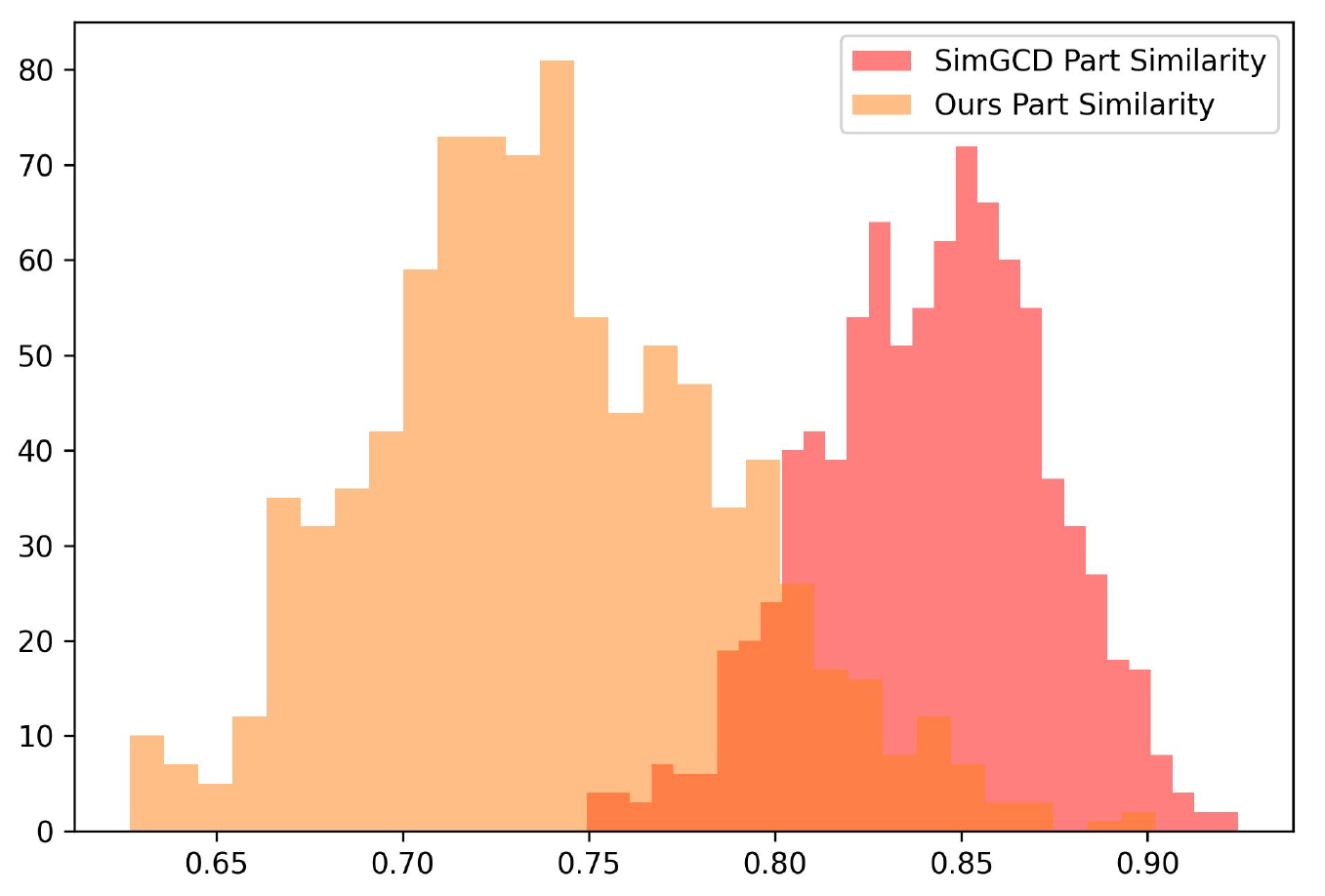}   
    \caption{The similarity distribution of discriminative parts between similar but distinct categories.}
    \label{fig:partdist}
    \vspace{-8mm}
\end{figure}

Additionally, We compare our APL's effectiveness on a range of generic image recognition datasets with other methods which focus on fine-grained gneralized category discovery. Table~\ref{table_generic} presents results on CIFAR-10, CIFAR-100 and ImageNet-100. Unlike fine-grained datasets, they require fewer fine-grained parts to distinguish between similar but different categories and share less knowledge across vastly different categories. Despite this, we consistently outperform all our baselines, with improvements of 3.4\%, 3.2\% and 0.7\% on novel categories of CIFAR-10, CIFAR-100 and ImageNet-100, respectively. We present the t-SNE visualization of our method in Figure~\ref{fig:parttsne} and the comparison of similarity distribution of parts in Figure~\ref{fig:partdist}.

\subsection{Ablation Study}
We employ SimGCD as the baseline and conduct ablation studies
to validate the effectiveness of different components in our APL. 
More ablations are shown in the Appendix.

\noindent\textbf{Part Discovery.} Table~\ref{table:ablation1} showcases the results of different approaches and various variants of our method for part discovery. 
(1) Compared to the baseline SimGCD, the decreased performance observed (row 2) when clustering patches from individual images to discover parts indicates that naively incorporating local information can introduce noise. Moreover, single-image clustering struggles with consistency and lacks robustness. 
(2) Directly treating our extracted part priors \( F_{\texttt{prior}} \),
defined in Equ~\ref{equ:priors}, as parts and using them to complement the \( \texttt{CLS} \) feature in SimGCD slightly improves its performance, demonstrating the positive impact of meaningful part information in GCD (row 3). 
(3) Moreover, we further introduce our learnable part queries to iteratively and dynamically query the DINO priors \( F_{\texttt{prior}} \) and image patches \( F_{\texttt{patch}} \) to address the consistency and correspondence challenges in GCD's part discovery (row 4). The significant performance improvements, i.e., 4.1\% on novel categories, demonstrate the robustness of using learnable queries to capture cross-image shared part semantics. 
(4) We enhance the soft assignment of learnable queries with image patch features \( F_{\texttt{patch}} \) from the previous model to a hard assignment based on the Gumbel-Softmax, defined in Equ~\ref{eq:4}, which led to further improvements of 1.6\%. 

\begin{table}[htbp]
\begin{center}
\small
\renewcommand{\arraystretch}{1.07}
\setlength{\tabcolsep}{0.125cm}
\begin{tabular}{l|lccc}
        \toprule[1pt]
        \multicolumn{1}{l}{} &{Method} &All&Known&Novel  \\
        \hline
        1&SimGCD                                &53.8 &71.9 &45.0       \\ 
        2&1+ $k$-Means      & 51.0 &  70.3  &  41.7     \\
        3&1+ DINO part prior                 & 55.1 & 73.9    &  46.0     \\
        4&3 + Learnable queries     &  58.5  & 74.7  & 50.1    \\
        5&4+ Hard assignment                 & 60.1  &  77.6   & 51.2    \\

        \bottomrule[1pt]
        \end{tabular}
\end{center}
\vspace{-5mm}
\caption{Ablation study of part discovery.} 
\vspace{-2mm}
\label{table:ablation1}
\end{table}
\begin{table}[htbp]
\begin{center}
\small
\renewcommand{\arraystretch}{1.07}
\setlength{\tabcolsep}{0.125cm}
\begin{tabular}{l|lccc}
        \toprule[1pt]
        \multicolumn{1}{l}{} &{Method} &All&Known&Novel  \\
        \hline
        1&SimGCD  + part discovery              &   55.1 & 72.5    & 46.7      \\ 
        2&1+ part contrastive loss                          &55.2   & 73.2  & 46.5    \\
        3&1+ min-max contrastive  loss                            &57.8  &75.7  &49.1     \\
        4&1+ diversity loss                            &  56.7  & 72.0  &  49.6   \\
        5&3+ diversity loss                           &  60.1 &  77.6  &  51.2 \\
        
        \bottomrule[1pt]
        \end{tabular}
\end{center}
\vspace{-5mm}
\caption{Ablation study of learning objectives.} 
\vspace{-3mm}
\label{table:ablation2}
\end{table}

\noindent\textbf{Part Learning Objective.}
Table~\ref{table:ablation2} presents the performance of our method under various part learning objectives. (1) The baseline, developed on the SimGCD framework, integrates our part discovery, specifically \( P_{\texttt{part}} \), as defined in Equ~\ref{eq:4}. However, it only incorporates the SimGCD loss \( \mathcal{L}_{GCD} \) from Equ~\ref{eq:gcd} and does not employ any of our part learning losses.
(2) Row 2 extends the baseline by implementing a common contrastive loss among parts, ensuring that all parts of positive samples are similar, whereas all parts of negative samples are dissimilar. The results indicate no improvement for novel categories but a slight enhancement in known categories, attributed to enhanced discriminability among fine-grained object parts across images.  
(3) When we substitute our all-min contrastive loss, we observe significant improvements of 2.7\%, 2.8\%, 2.4\% in the all, known, and novel categories, respectively.
It adaptively identifies the most discriminative part in negative samples to boost discriminability, while permitting similarity in other parts to enhance the transferability of shared parts across different categories. 
(4) We also evaluate the independent effect of diversity loss on our baseline, showing that emphasizing the distinctiveness of various parts within an image effectively enhances the discriminability of part-level features. 
(5) Finally, we validate the combined effects of the two part learning losses.
\section{Conclusion}
This paper underscores the trade-off between discriminability and generalization in existing methods and introduces APL, which enhances both aspects simultaneously and seamlessly integrates into GCD frameworks. APL utilizes DINO prior and introduces novel all-min contrastive and diversity loss functions to enhance knowledge transfer, resulting in superior performance across benchmarks. 
\textbf{Acknowledgment:} 
This work was supported by the National Natural Science Foundation of China (No.62206174).
{
    \small
    \bibliographystyle{ieeenat_fullname}
    \bibliography{main}
}

\end{document}